%% file: main_acl.tex
\documentclass[11pt]{article}

\IfFileExists{acl.sty}{%
  \usepackage[preprint]{acl}%
  \newcommand{\aclofficialstyle}{}
}{%
  \usepackage[margin=1in]{geometry}
}

\usepackage{times}
\usepackage[T1]{fontenc}
\usepackage[utf8]{inputenc}
\usepackage{microtype}
\usepackage{amsmath}
\usepackage{graphicx}
\usepackage{booktabs}
\usepackage{multirow}
\usepackage{xcolor}
\usepackage{natbib}

\newcommand{\dsm}{DeepSeek-V4-Flash}
\newcommand{\glm}{GLM-5.3-Flash}
\newcommand{\gpt}{GPT-6.1-Sol}

\title{From Unreliable to Usable: Decomposing Failure Modes and Confidence\\
Signals in LLM Entity Extraction --- Evidence from Cyber Threat Intelligence}

\author{Jiahong Qian \\
  Shanghai Jian Qiao University, Shanghai, China \\
}

\begin{document}
\maketitle

\input{sections/00_abstract}
\input{sections/01_intro}
\input{sections/02_related}
\input{sections/03_setup}
\input{sections/04_findings}
\input{sections/06_conclusion}
\input{sections/05_limitations}

\ifdefined\aclofficialstyle\else
  \bibliographystyle{plainnat}%
\fi
\begingroup
\raggedright
\bibliography{references_en}
\endgroup

\appendix
\input{sections/appendix}

\end{document}

%% file: sections/00_abstract.tex
\begin{abstract}
Audits of large language model (LLM) extraction from full-length documents report severe unreliability, leaving two questions open: what the failures actually consist of, and whether confidence signals can turn unreliable output into usable output. We run a preregistered, fully frozen evaluation of two new-generation low-cost LLMs (\dsm{}, \glm{}) with a flagship-tier control (\gpt{}) on a public corpus of 354 incident campaigns (315 unique reports, 630 model--report evaluations), scoring extraction quality against gold, confidence calibration, and robustness to single-sentence perturbations. Three findings emerge. (1)~Unreliability persists across tiers: micro precision is 0.03--0.34 and verbalized calibration error (ECE) 0.53--0.94 across both splits; the gap does not close at the flagship tier. (2)~The dominant failure is not hallucination: a full mention census plus a three-tier annotation of 360 sampled candidates (two independent LLM annotators from vendors outside the evaluated set, Cohen's $\kappa=0.79$ on the preregistered acceptance line, fully disclosed) back-projects to 65.6\% background mentions and 19.9\% gold-scope gaps, with only 1.2\% unsupported candidates. (3)~Confidence ranks but does not verify: zero-cost self-consistency-frequency tiering passes a preregistered top-30\% precision-gain bar in all six test-split model--type cells (4/6 on train), the zero-cost text-frequency$\times$SC combination wins 3, ties 2, loses 1 across cells, yet relation-level counterfactuals on all 45 targets show that a single misattribution sentence flips high-confidence judgments in 95.6--100\% of cases, and even a background sentence explicitly marked as unrelated still binds a fake vulnerability in up to 26.7\% of them. We conclude with a deployment boundary: confidence tiering may set review priorities, never truth.
\end{abstract}

%% file: sections/01_intro.tex
\section{Introduction}
\label{sec:intro}

When large language models (LLMs) fail at information extraction, the failures are commonly attributed to hallucination. Auditing 315 unique real-world documents (354 incident campaigns; 630 model--report evaluation runs) under a frozen, preregistered protocol with three-tier verified annotation, we find hallucination accounts for only 1.2\% of false positives --- 65.6\% are \emph{background mentions}: correct strings extracted from the wrong context. We then ask whether confidence signals can still make such a system usable, and show they \emph{rank} but do not \emph{verify}: across all 45 counterfactual targets, a single misattribution sentence flips high-confidence judgments in 95.6--100\% of cases, and one background sentence explicitly marked as unrelated still binds a fake vulnerability in up to 26.7\%.

These results address a question that outlives any single domain, and the stakes are concrete. LLM-based extraction pipelines for structured threat intelligence claimed 83--97\% precision \citep{chen2024survey}; a critical audit \citep{mezzi2025unreliable} traced those numbers to toy inputs of 18--106 words and showed that at real scale recall falls below 0.80 and calibration worsens; a reproducibility audit found only 5 of 12 extraction systems re-runnable under their published protocols \citep{hakim2026matcher}. But ``unreliable'' is a verdict, not a diagnosis --- and verbalized confidence is itself severely miscalibrated \citep{xiong2024uncertainty,guo2017calibration}. It does not say \emph{which part} of the pipeline is broken --- the model's faithfulness, the evaluation scope, or the task definition --- and it does not say whether the residual uncertainty can be managed. Two questions follow. First, \textbf{what is the failure made of?} Hallucinated content, wrong context, or a mismatch between what the model extracts and what the annotation counts as correct demand different repairs. Second, \textbf{can confidence make unreliability usable?} If high-confidence outputs can be prioritized for scarce human review, an unreliable extractor may still be deployable --- unless confidence itself is untrustworthy.

We answer both questions on a testbed chosen for measurement quality rather than topical breadth: cyber threat intelligence (CTI) reports, long documents describing intrusion campaigns, from which models must extract three entity types (threat-actor groups, vulnerability identifiers, attack techniques). The domain offers three assets that generic extraction benchmarks rarely combine: a public corpus with gold annotations for 354 campaigns \citep{dimitizio2022software}, a published critical audit whose protocol and splits we reuse so results are directly comparable \citep{mezzi2025unreliable}, and a real deployment need --- security analysts already triage extracted indicators by confidence, so ``usable'' has an operational meaning. Our evaluation protocol is explicit and frozen end to end: multiset entity matching with alias equivalence classes, dual confidence proxies (verbalized self-report and self-consistency frequency under 5-sample resampling), preregistered acceptance thresholds, relation-level counterfactual perturbations, and a three-tier annotation study (textual mention / event relevance / relation correctness) on 360 stratified candidates, labeled by two independent LLM annotators from vendors outside the evaluated set with adjudication fully disclosed ($\kappa=0.79$ on the borderline-excluded preregistered line).

\paragraph{Findings.} (1)~\emph{Unreliability persists across model tiers.} Micro precision spans 0.03--0.34 against recall of 0.81--0.99; verbalized ECE is 0.53--0.94 across both splits, and a full-scope control at the flagship tier (\gpt{}) shows the same pattern (0.63--0.94). (2)~\emph{The dominant failure is background contamination, not hallucination.} 99.7--99.8\% of false positives have an explicit textual source in the report; weighted back-projection of the three-tier annotation yields 65.6\% background mentions and 19.9\% gold-scope gaps against 1.2\% unsupported candidates. (3)~\emph{Confidence signals decompose into separable, entity-type-dependent properties.} SC frequency passes a preregistered ranking bar in all six test-split model--type cells (4/6 on train, with both vulnerability cells just under the line); controlling for text frequency, its partial correlation with correctness is significant in 6/6 groups ($r=0.30$--$0.43$); mention strength tracks event centrality for actors while sampling stability tracks correctness for vulnerabilities and techniques. (4)~\emph{No confidence signal measures truth.} A single misattribution sentence flips high-confidence judgments in 95.6--100\% of 45 counterfactual targets; verification signals --- self-run or third-party --- inherit the same blindness; repairs only help when matched to the failure mode, ranking gains transfer to a newer corpus, and thresholds must be recalibrated per corpus.

\paragraph{Contributions.} This is a systematic evaluation study: model choices and prompting configurations are experimental settings, ranking heuristics are baselines, error analysis and counterfactuals are the core explanation, and repair experiments serve the boundary characterization. The contributions are three:
\begin{itemize}
\item \textbf{A reproducible reliability measurement.} To our knowledge (as of 2026-10, within audited coverage), the first evaluation framework merging multiset PRF with alias equivalence classes, dual-proxy calibration, three-tier evidence annotation, and preregistered thresholds into one explicit, frozen protocol --- executed on the public audit corpus as a direct response to the reproducibility audit above. All protocol, code, and frozen evaluation outputs accompany this preprint and will be made public; an anonymized copy is additionally available to reviewers as supplementary material.
\item \textbf{A quantified failure-mode decomposition.} A full-scope mention census plus a preregistered three-tier annotation study locating 85.5\% of the false-positive mass in background mentions and gold-scope gaps, against 1.2\% hallucination --- reframing ``LLMs hallucinate entities'' as ``LLMs over-attend to context''.
\item \textbf{A tested deployment boundary.} Evidence across confidence signals from self-report to third-party verifiers that ranking benefit is real, signal--type matching matters, calibration scope matters, and no tested signal survives relation-level counterfactuals; ranking gain transfers to a newer corpus while thresholds must be recalibrated.
\end{itemize}

%% file: sections/02_related.tex
\section{Related Work}
\label{sec:related}

\paragraph{Calibration and verbalized uncertainty.}
Classical calibration is measured by ECE and reliability curves \citep{guo2017calibration}; for LLMs, verbalized confidence is often miscalibrated \citep{xiong2024uncertainty,tian2023justask}, sampling consistency carries partial signal \citep{wang2023selfconsistency,kadavath2022know}, and selective prediction formalizes the risk--coverage trade-off our tiering analysis instantiates \citep{geifman2017selective}. We add a three-way distinction --- output \emph{consistency}, \emph{ranking benefit}, and \emph{probabilistic calibration} are different quantities that must not be conflated --- and test all three against counterfactual ground truth.

\paragraph{LLM-based extraction from threat intelligence.}
Structured extraction from CTI reports attracted rapid LLM adoption \citep{llmtikg2024,mouiche2026ctikg,bouchiha2026tatickg} alongside benchmarks \citep{ctibench2024,sevenllm2024,alam2025athenabench,attackclass2026,tracecti2026} and corpora \citep{dimitizio2022software,cyberner2025,stix2026dataset}. A critical audit \citep{mezzi2025unreliable} overturned headline extraction numbers by exposing scoring loopholes and finding LLMs unreliable with overconfidence; follow-ups target pipeline dependencies \citep{benchmark2026pipeline}, automated TTP and strategic-intelligence extraction \citep{cheng2026ttprint,echchammakhy2026stiner}, agentic KG construction \citep{li2026beacon}, calibrated verification services \citep{vasan2026sok,tang2025polar}, and corpus fragmentation \citep{ctiecho2026}. Reproducibility audits question KG-construction pipelines \citep{dinis2025ragrecon,hakim2026matcher}, and systems such as CTINexus \citep{cheng2025ctinexus} have been re-examined under stricter scoring; \citet{chen2024survey} systematizes the application landscape. Our work is a systematic evaluation study on the audit's own public corpus: we reuse its splits for comparability, and extend the audit line from ``unreliable'' to a measured, preregistered account of what the unreliability consists of and where confidence helps.

\paragraph{Failure-mode analysis of extraction.}
Prior CTI work reports misses, over-extraction, and misbinding qualitatively \citep{alam2022cyner,wang2025threatlevel,mezzi2025uncertainty,heo2026whentrust}. We quantitatively decompose false positives into mention-level and event-level layers with a preregistered three-tier annotation, and separate hallucination (1.2\%) from background contamination (65.6\%) and gold-scope gaps (19.9\%).

%% file: sections/03_setup.tex
\section{Experimental Setup}
\label{sec:setup}

\subsection{Task, corpus, and models}
\label{sec:task}

Given a full-length CTI report, the model extracts three entity types: \textbf{threat-actor groups} (APT), \textbf{vulnerability identifiers} (CVE), and \textbf{attack techniques} (AV, MITRE ATT\&CK-style names \citep{strom2020attack}). We use the public Di Tizio corpus \citep{dimitizio2022software} under the exact campaign-level 70/30 split of the critical audit \citep{mezzi2025unreliable}: 354 campaigns (248 train / 106 test at campaign level), 315 train and 93 test reports per model --- the same report split reused for both models, for 630 model--report runs in total; Table~\ref{tab:main} reports the test split. Gold follows the corpus's STIX 2.1 structure \citep{oasis2021stix} and marks \emph{primary} entities per campaign. We evaluate two new-generation low-cost models (\dsm{}, \glm{} --- the cheapest ``flash'' service tiers of their vendors) as subjects and \gpt{} as a flagship-tier full-scope control; all runs use frozen prompts, segmentation (cap 40k tokens, overlap 800), temperature-0 extraction with 5-sample self-consistency (SC) resampling (T$=$0.7) per segment. Every reported number is regenerated from frozen batch outputs; the pipeline records gateway and model identity per call (6,048 used segment-level calls on the main table; full call/token accounting in the appendix).

\subsection{Scoring and confidence signals}
\label{sec:protocol}

\paragraph{Multiset matching with alias equivalence classes.}
Predictions and gold are matched as multisets under type-specific normalization: alias equivalence classes with word/digit boundaries (APT), fault-tolerant CVE regexes covering PDF mojibake separators and legacy CAN- prefixes, and a conservative 13-value AV synonym table. This counts every mention as a prediction while gold lists primary entities --- a deliberately strict-on-the-model scope whose FP semantics is disambiguated by the annotation study (\S\ref{sec:failures}).

\paragraph{Confidence signals.}
(1)~\emph{Verbalized}: the model's self-reported 0--100 confidence; (2)~\emph{SC frequency} (\texttt{sc\_freq}): the share of 5 resamples reproducing the entity --- a zero-extra-cost measure of output consistency; (3)~\emph{text frequency}: the count of the entity string in the report, a black-box signal requiring no logits. Their rank-percentile mean forms the zero-cost combination \texttt{textfreq\_sc}; two external verifiers (self-verification and a third-party calibrated decision model) are evaluated against these in \S\ref{sec:signals}.

\paragraph{Preregistered acceptance.}
Before evaluation we registered: a tiering bar (top-30\% confidence-bucket precision $\geq$ overall $+0.10$), an annotation agreement line ($\kappa \geq 0.7$), and a hypothesis register (H1--H6) separating measured claims from pending ones. Failing a registered bar must be restated as a negative result.

\subsection{Three-tier annotation of sampled candidates}
\label{sec:annotation}

To separate \emph{what the model extracted} from \emph{what the annotation counts}, all evaluation units of the test split were partitioned by a deliberately lenient mention detector into TP / FP-mentioned / FP-unsupported, and a 360-row stratified subsample (60 per model$\times$type; the FP-unsupported stratum fully sampled, the rest balanced across SC-frequency tertiles) was labeled on three tiers: \textbf{L1} textual mention (does the string appear), \textbf{L2} event relevance (does it belong to this incident, not background or related work), \textbf{L3} relation correctness (is the binding and attribution right). The second-annotator role was filled by two independent LLMs (kimi-k3, minimax-m3; \emph{neither vendor is among the evaluated models}) under a frozen prompt with temperature 0 and golden-item gating; 36 hard disagreements went to a third adjudicator model and 27 borderline items to a conservative rule resolving toward the model-unfavorable side. Agreement is $\kappa_{L2}=0.68$ three-class, \textbf{0.79 excluding borderline} (meeting the preregistered 0.7 line), $\kappa_{L3}=0.74$. Stratified sampling weights back-project row-level statistics to the full FP composition (\S\ref{sec:failures}). The mechanism is disclosed as LLM-assisted dual annotation throughout --- never as two human annotators.

%% file: sections/04_findings.tex
\section{Results}
\label{sec:results}

\subsection{Finding 1: extraction remains unreliable at every tier}
\label{sec:unreliable}

\begin{table*}[t]
\centering
\small
\begin{tabular}{llccccc}
\toprule
Model & Type & Micro P & Micro R & Micro F1 & Macro F1 & 95\% CI (macro F1) \\
\midrule
\glm{} (test, 93) & APT & 0.284 & 0.978 & 0.441 & \textbf{0.645} & [0.576, 0.714] \\
 & CVE & 0.069 & 0.884 & 0.129 & \textbf{0.225} & [0.156, 0.295] \\
 & AV & 0.174 & 0.806 & 0.287 & \textbf{0.309} & [0.248, 0.373] \\
\dsm{} (test, 93) & APT & 0.336 & 0.957 & 0.497 & \textbf{0.675} & [0.602, 0.743] \\
 & CVE & 0.062 & 0.855 & 0.115 & \textbf{0.224} & [0.154, 0.292] \\
 & AV & 0.297 & 0.806 & 0.434 & \textbf{0.391} & [0.314, 0.469] \\
\bottomrule
\end{tabular}
\caption{Multiset entity-extraction PRF, test split (train-split rows are analogous; appendix). High recall with low precision is structural: the models find most gold entities but drown them in extra candidates (\glm{} test emits 15.1 CVE strings per report on average against at most 2 gold primary CVEs).}
\label{tab:main}
\end{table*}

Table~\ref{tab:main} shows the structural signature: recall across types is 0.81--0.99 while micro precision is 0.06--0.34. The models ``find'' most gold entities --- primary-entity coverage is 0.60--0.95 --- but bury them under $10$--$20\times$ extra strings. Calibration is worse: mean verbalized confidence sits at 80--90+ against unit-level accuracy of 0.05--0.32, and test-split verbalized ECE reaches 0.59--0.88 (CVE: 0.84--0.94), more severe than the critical audit's zero-shot scope \citep{mezzi2025unreliable}. The flagship control replicates the pattern (micro P 0.057 / R 0.899 on CVE; ECE 0.63--0.94; APT macro-F1 0.684 statistically indistinguishable from \dsm{}'s 0.675): \textbf{upgrading the model tier does not close the reliability gap}. SC frequency, by contrast, is a markedly better probabilistic-calibration proxy at zero extra cost (CVE ECE 0.16--0.21 vs verbalized 0.84--0.94); on risk--coverage (AURC), however, the advantage narrows to 4/6 on the \glm{} side --- ranking benefit and probabilistic calibration are separable properties (\S\ref{sec:signals}). The critical audit's higher CVE precision (0.67--0.72) comes from a stricter primary-only matching scope; differently-scoped numbers are never placed in one table.

\subsection{Finding 2: failures are background contamination, not hallucination}
\label{sec:failures}

A full-scope mention census over all evaluation units --- type-adaptive, deliberately lenient detection so that ``unsupported'' is a lower bound --- shows that \textbf{99.7\% (\dsm{}) / 99.8\% (\glm{}) / 99.7\% (\gpt{})} of CVE false positives have an explicit textual source in the report; unsupported output is at most 0.2--11.2\%. The three-tier annotation (Table~\ref{tab:threetier}) turns this census into an event-relevance-layer quantification: back-projected by sampling weights, \textbf{85.5\% of the full FP composition (across all three entity types) is background mentions (65.6\%) plus gold-scope gaps (19.9\%)} --- entities the report genuinely mentions and that are event-relevant, but which fall outside the primary-entity gold set --- while unsupported phantom candidates are only \textbf{1.2\%} (the unweighted 10.3\% is inflated by the fully-sampled stratum). Confidence separates the FP mass by nature: rows in the high SC-tertile are event-relevant in 62\% of cases (mostly scope mismatch --- the model got the event right, gold does not include it) versus 26\% in the low tertile, while 23\% of high-confidence gold rows are judged not event-relevant --- the scope gap is bidirectional (gold itself has boundaries). The same decomposition holds on a newer 2022--2025 corpus (annotation-derived gold, $\kappa=0.53$ borderline-excluded --- Limitations), where relation-level errors (\texttt{L3=no}) are entirely absent and FPs are again dominated by background mentions (51.2\%). Relation-level KG alignment completes this picture: aligning extracted relation triples against the gold graph yields F1 of only 0.05--0.33 (flagship CVE P 0.049 / R 0.750 / F1 0.092 --- precision below both flash tiers with recall on par), because one extra entity is one FP at the entity level but a spurious edge at the graph level --- \textbf{graph quality must be scored at the relation level}, not by entity counts or visual inspection.

\begin{table}[t]
\centering
\small
\begin{tabular}{lrrr}
\toprule
Failure mode & $n$ & $n$\% & w\% \\
\midrule
Background mention (L2$=$no) & 134 & 37.2 & \textbf{65.6} \\
Event-relevant, not in gold & 120 & 33.3 & \textbf{19.9} \\
TP (relevant, bound, in gold) & 52 & 14.4 & 10.0 \\
No textual support (lower bound) & 37 & 10.3 & \textbf{1.2} \\
Gold-matched, not event-relevant & 11 & 3.1 & 2.7 \\
Unresolved borderline & 5 & 1.4 & 0.4 \\
Event-relevant, relation wrong & 1 & 0.3 & 0.2 \\
\bottomrule
\end{tabular}
\caption{Three-tier failure-mode distribution on the 360-row stratified subsample, all three entity types (w\% = back-projected to the full FP composition via sampling weights; the top rows are FP-composition shares). Relation misbinding is a 1-row residual in natural output.}
\label{tab:threetier}
\end{table}

The reading is blunt: the reliability problem of modern LLM extraction on long documents is \emph{not} a faithfulness problem. The models almost never invent entities; they report strings whose truth they cannot adjudicate against \emph{this} event --- ``mentioned'' is mistaken for ``used''. Repairs and deployments should therefore target context disambiguation and review prioritization, not factuality.

\subsection{Evidence faithfulness: outputs are weakly anchored to their own evidence}
\label{sec:ehr}

Beyond entity strings, the protocol scores a three-tier evidence hit rate (EHR): whether the segment carrying an extraction actually contains the evidence it claims, at graded evidence-similarity thresholds. EHR is low overall (macro 0.19--0.51 across models, splits, and thresholds: test-split @0.95 = 0.235 \glm{} / 0.371 \dsm{} / 0.396 \gpt{}), with \dsm{}'s evidence sentences leading \glm{}'s by about $+0.14$ --- the evidence-capability gap exceeds the extraction-capability gap, and even the flagship stays below 0.5. The cause is a value-vs-sentence misalignment (models emit canonical technique names where reports use natural-language descriptions; decomposition in the appendix), and the scope is verbatim-anchored, stricter than and incomparable to RAGAS-style faithfulness --- \textbf{explanation reliability is the current weak spot of LLM extraction}, independent of entity precision.

\begin{figure}[t]
\centering
\includegraphics[width=\columnwidth]{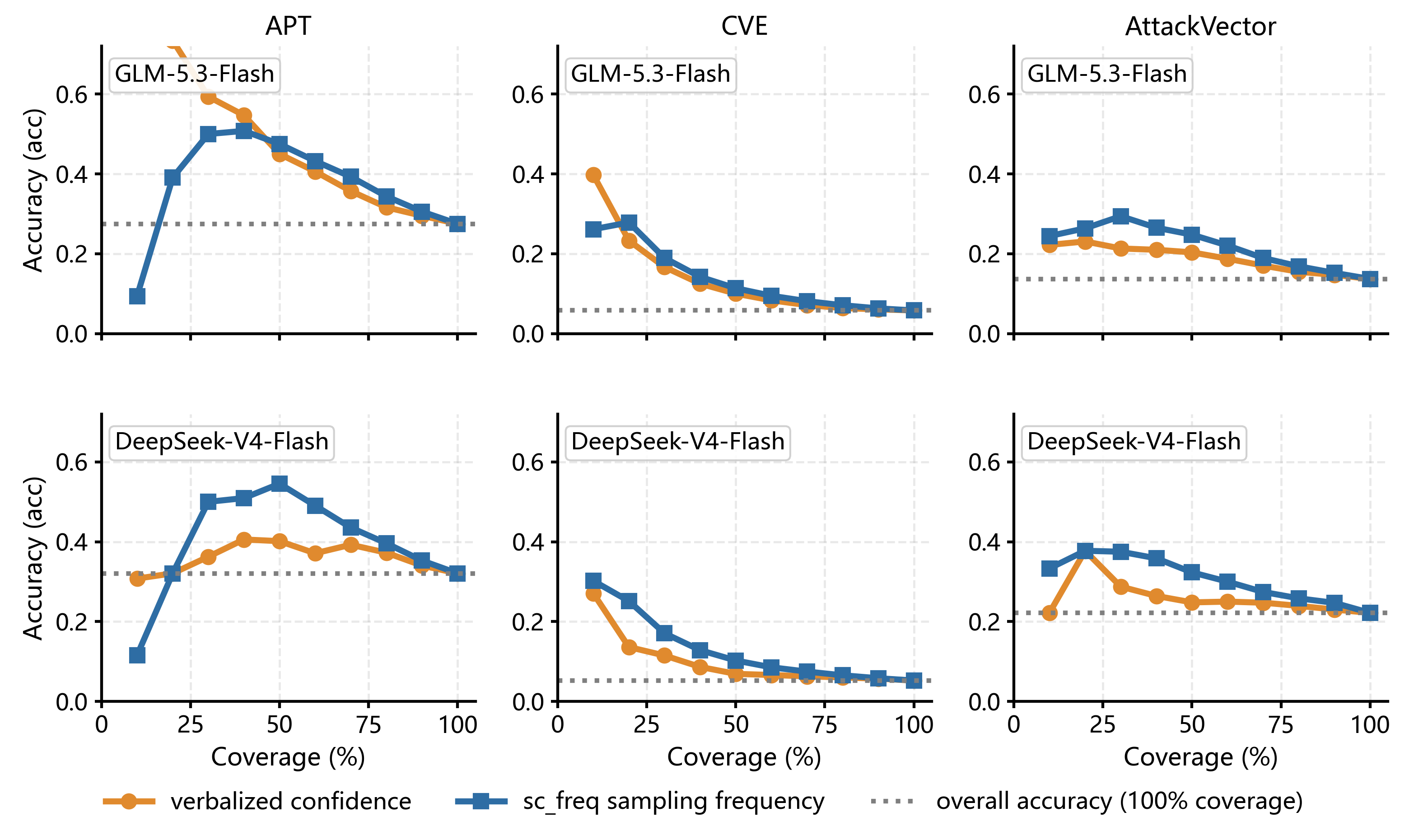}
\caption{Coverage--accuracy curves of confidence tiering (test split): rows are \glm{} and \dsm{}; columns are APT / CVE / AV. Orange circles: verbalized self-report; blue squares: SC frequency; gray dotted: overall accuracy. SC-frequency tiering is steep and monotone (e.g.\ \dsm{} APT precision 0.500 at 30\% coverage vs overall 0.321); verbalized is nearly flat (97\%+ of self-reports pile into 80--100).}
\label{fig:coverage}
\end{figure}

\subsection{Finding 3: confidence signals decompose --- and rank, not verify}
\label{sec:signals}

\begin{table*}[!t]
\centering
\small
\setlength{\tabcolsep}{3.5pt}
\begin{tabular}{llcccc}
\toprule
Model & Type & rand. & text\_freq & verb. & sc\_freq \\
\midrule
\dsm{} & APT & $-.004{\pm}.045$ & \textbf{+0.367} & +0.042 & +0.179 \\
 & CVE & $\pm$.011 & +0.084 & +0.063 & +0.118 \\
 & AV & $\pm$.037 & $-$0.034 & +0.066 & \textbf{+0.153} \\
\glm{} & APT & $\pm$.039 & +0.340 & +0.319 & +0.225 \\
 & CVE & $\pm$.013 & +0.082 & +0.109 & \textbf{+0.131} \\
 & AV & $\pm$.024 & $-$0.063 & +0.076 & \textbf{+0.157} \\
\bottomrule
\end{tabular}
\caption{Unified-candidate-set ranking baselines, test split: top-30\% precision gain against the preregistered $+0.10$ bar (random = 200 seeds $\pm$ sd). The zero-cost combination \texttt{textfreq\_sc} (omitted column; Fig.~\ref{fig:signals}) wins 3, ties 2, loses 1 of 6 cells; the flagship tier confirms the mechanism at \texttt{textfreq\_sc} $+0.454$ on APT, the largest value across all 9 combinations.}
\label{tab:signals}
\end{table*}

\begin{figure}[t]
\centering
\includegraphics[width=\columnwidth]{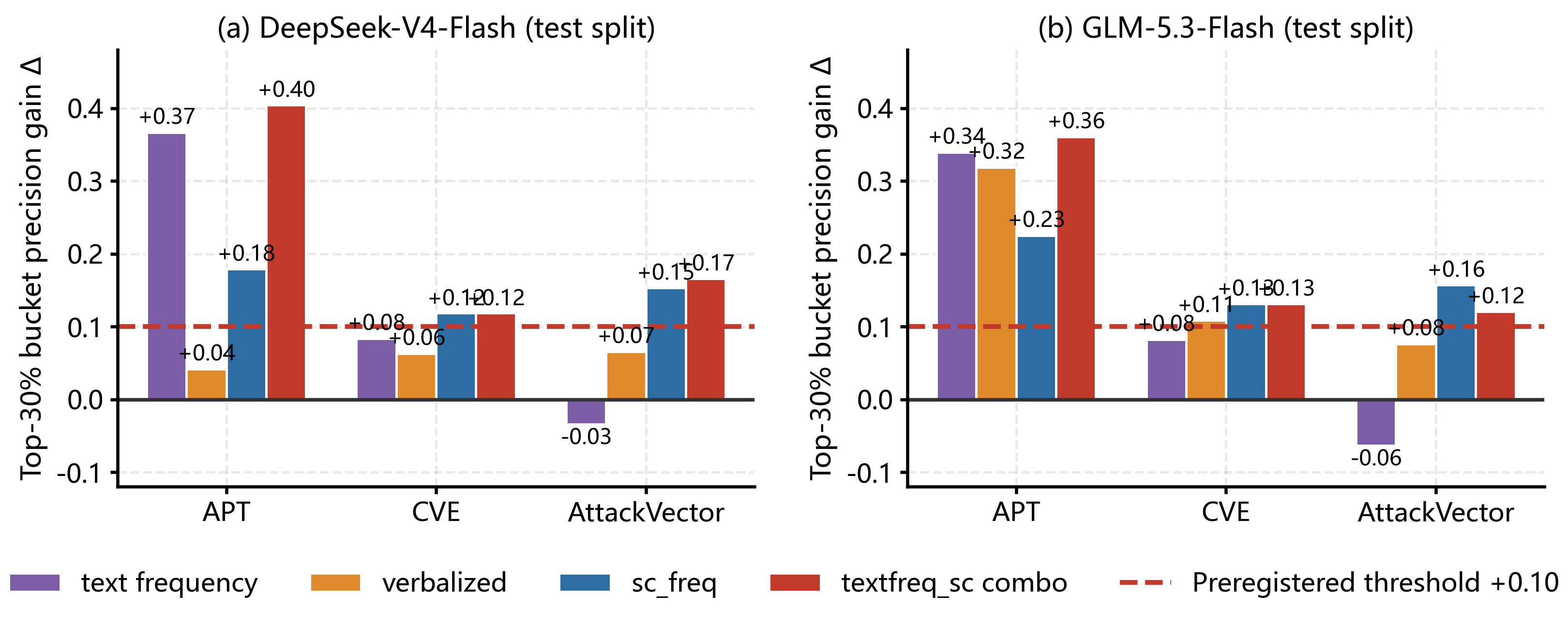}
\caption{Ranking baselines including the zero-cost combination \texttt{textfreq\_sc}: on actor entities mention strength measures event centrality; on techniques text frequency is negative while SC works. Red dashed line: preregistered $+0.10$ bar.}
\label{fig:signals}
\end{figure}

Three separable properties, three answers. \emph{Ranking benefit is real and signal--type matched} (Table~\ref{tab:signals}, Fig.~\ref{fig:signals}): SC frequency passes the preregistered bar in all six cells; the zero-cost \texttt{textfreq\_sc} combination is the best overall choice (best or tied in 5/6 cells), but text frequency alone \emph{hurts} on attack techniques (canonical names decoupled from surface phrasing) while SC works --- ``which confidence proxy is better'' is an engineering decision over signal$\times$type, not a binary. \emph{SC is not a proxy for text frequency}: controlling for $\log(1{+}\text{text\_freq})$, the partial correlation of SC frequency with correctness is significant in 6/6 groups ($r=0.30$--$0.43$, $p\leq 0.0005$), while \dsm{}'s verbalized confidence is fully absorbed (partial $r=0.000$, $p=0.997$) --- self-report is nearly useless as a ranking signal except where it carries event-centrality information (\glm{}-APT, $r=0.187$). \emph{Scope changes calibration verdicts}: recomputed on the adjudicated three-tier labels, SC-frequency calibration error \emph{shrinks} in 5/6 groups (e.g.\ 0.62$\to$0.38), because scope-mismatch FPs charged against the main-table scope are recovered as correct event-centrality rankings --- verbalized overconfidence, by contrast, is absolute and unaffected. \emph{What a review budget buys is type-specific}: simulating fixed review budgets in SC-frequency order, a 20\% budget on vulnerabilities locks in 96\% of correct items (with the head's selective risk disclosed below), while the actor ranking's top 5--10\% is \emph{enriched with errors} (11 of the first 13 units are misbinding FPs, density gain negative until a 30\% budget) --- \textbf{review-from-the-top fails on actors}, and fixed thresholds must be set per type (full budget--benefit tables in the appendix).

\subsection{Finding 4: none of it verifies truth}
\label{sec:counterfactual}

\begin{table*}[!t]
\centering
\small
\setlength{\tabcolsep}{3pt}
\begin{tabular}{llcc}
\toprule
Perturbation & Outcome & \dsm{} & \glm{} \\
\midrule
neg (retraction) & true CVE abandoned & 34.1\% & \textbf{76.2\%} \\
bg (unrelated) & fake CVE bound to event & \textbf{26.7\%} & 4.5\% \\
mis (attribution) & injected CVE bound & 95.6\% & \textbf{100\%} \\
mis (attribution) & injected org.\ accepted & 73.3\% & \textbf{95.6\%} \\
\midrule
\multicolumn{2}{l}{Control: assertion (v1), fake CVE bound} & 86.7\% & 97.8\% \\
\bottomrule
\end{tabular}
\caption{Relation-level counterfactuals: one injected sentence per report (all 45 targets, both models). Retained high confidence after binding: verbalized 58--98, SC frequency 0.50--0.90.}
\label{tab:cf}
\end{table*}

If confidence measured truth, sentences that change the facts should change the scores. They do not (flash tier; the flagship counterfactual batch was not run, see Limitations). All perturbations are synthetic by construction --- injected assertions contradict the facts, gold never updates, and nothing is extrapolated to real retraction or attribution-update streams. Injecting one style-matched sentence per report (Table~\ref{tab:cf}): an attribution assertion makes both models rebind the event to an injected organization at high confidence (73.3\% / 95.6\%); even a sentence explicitly flagged as \emph{unrelated to this case} still binds a fake vulnerability in 26.7\% (\dsm{}) of targets, and every residual binding carries above-average confidence (verbalized 98.1 / SC 0.774). Symmetrically, one retraction sentence erases the \emph{true} vulnerability from 34.1\% / 76.2\% of outputs while majority sampling still reproduces it --- true and fake entities are symmetric in their exposure to surface text. The fake entity's mean self-reported confidence (98.0 / 85.6) is indistinguishable from the true entity's, and its SC frequency is identical across models (0.892): \textbf{SC frequency measures textual consistency, not fact}. Self-verification does not rescue this: \dsm{}'s verify signal ranks at $+0.154$ on actors, its ``event-related'' rate on mentioned FPs (92.5\%) matching true entities (93.3\%) --- it mistakes mention for relevance --- while \glm{}'s verifier retains discrimination (87.1\% vs.\ 50.1\%; verify alone $+0.371$, SC$\times$verify $+0.413$, the best actor ranker of all signals), and a third-party calibrated decision model (TypeSafe Jev; 1,461 calls, \$0.051; protocol and per-call logs in the appendix) shows the same law: its increment depends on whether its window-level judgment semantics align with report-level gold. Across all tested signals --- self-report, sampling consistency, self-verification, third-party verification --- \textbf{none survives relation-level counterfactuals}: confidence sets review priorities; it must never adjudicate truth, and a verification component earns its place only when the verifier passes a one-shot FP-mentioned event-relevance precheck.

\subsection{Repair and generalization boundaries}
\label{sec:repair}

Retrieval-augmented and chain-of-thought repair helps actors (up to $+0.133$ macro-F1) and never helps vulnerabilities ($\Delta \leq \pm 0.011$, corroborated by an independent dual-gold recheck on a separate CVE corpus \citep{circl2024attack}); primary-entity schema alignment repairs actor over-extraction only ($+0.111$ and $+0.167$) --- repairs work when matched to the failure mode, consistent with Finding~2, and gains do not transfer across models (CoT flips sign on attack techniques). On an unfamiliar 2022--2025 corpus (30 group-deduplicated reports, three-tier annotation-derived gold), ranking-gain transfer is confirmed (top-30\% gains positive in 6/6 cells, never below the main pool; AURC improves across the board), while fixed thresholds do not transfer: after per-corpus quantile recalibration (unlabeled percentile cut points only), threshold buckets transfer fully for attack techniques (all recalibrated gaps $\leq 0.09$) and, for actors and vulnerabilities, cross-corpus differences are corpus-difficulty-driven with a conservative bias --- \textbf{threshold recalibration is a mandatory deployment step, not an option}. Cross-model ranking agreement persists on the new corpus (Spearman 0.63--0.77), and a top-10 report-level audit finds the top SC bucket carries analyst value (8/10 TTP-rich, none low) with no per-report guarantee. The deployable workflow follows: triage in descending \texttt{textfreq\_sc} order with per-type budgets --- a 20\% focused review zone plus tail sampling for vulnerabilities (bucket precision 0.17--0.19 vs overall 0.05--0.06), mandatory human review of the actor head (0.64--0.73 vs overall 0.27--0.32) --- protocol reproducibility belongs on review checklists for extraction papers, and tiering prioritizes governance queues while evidence checks gate ingested claims. These are review-simulation figures, not measured human effort (Limitations).

%% file: sections/06_conclusion.tex
\section{Conclusion}
\label{sec:conclusion}

We turned the ``unreliable'' verdict on LLM entity extraction into a measured, preregistered account: the dominant failure is background contamination (65.6\%) and gold-scope gaps (19.9\%), not hallucination (1.2\%); confidence signals are real but decompose into consistency, ranking, and calibration --- each useful, each type-dependent, none truth-tracking under flash-tier relation-level counterfactuals. The deployable boundary is correspondingly concrete: match the ranking signal to the entity type, recalibrate thresholds per corpus, and treat confidence tiering as review prioritization with humans or independent verifiers retaining truth. All protocol, code, frozen outputs, and annotation artifacts accompany this preprint and will be made public; an anonymized copy is additionally available to reviewers.

%% file: sections/05_limitations.tex
\section{Limitations}
\label{sec:limitations}

\paragraph{Domain scope of the testbed.} All absolute numbers come from one domain (threat-intelligence reports) and one gold-annotation convention (primary entities per campaign). We argue the \emph{mechanisms} (background contamination, signal--type matching, ranks-not-verifies) generalize, and corroborate direction on a newer corpus with independent annotation-derived gold; but cross-domain replication (biomedical, legal, scientific IE) is future work, and absolute precision values are not comparable across annotation scopes.

\paragraph{Measurement scope.} The serving gateways strip log-probabilities, so calibration proxies are signals that require no logits (verbalized + SC frequency), per the preregistered fallback plan; logprob-scope calibration awaits a direct-API environment. Gold covers three entity types --- campaign attribution is unscored (free-text names without comparable ground truth). Campaign-to-report alignment covers 88.4\% (18 unresolved cross-case surveys and 39 third-party group-level reports remain); no reported number depends on any single alignment, and sensitivity is recomputable from the open scripts. During one ablation round, 15 campaigns were blocked by a vendor content policy on a single channel (figures reported on the joint 84-report scope) --- CTI content itself can trigger LLM content policies, a production constraint on direct deployment.

\paragraph{LLM-assisted annotation.} The three-tier labels are produced by two LLM annotators from vendors outside the evaluated set plus a third adjudicator model --- disclosed as such throughout, never as two human annotators. Agreement meets the preregistered line ($\kappa=0.79$ borderline-excluded) but LLM annotation carries unknown failure modes relative to expert humans; the conservative adjudication rule resolves ambiguity toward the model-unfavorable side, and the generalization-corpus gold (lower $\kappa$, 0.53 borderline-excluded) bounds the confidence of that extrapolation. A human spot-check of a subsample is planned for the revision.

\paragraph{Model-tier coverage.} Subjects are two new-generation low-cost (``flash''-tier) models; the flagship control covers the main table, calibration, and mention census, but not the counterfactual, verification-signal, or generalization batches (cost-prohibitive at flagship rates; scope notes mark every affected table). Conclusions about verification signals are therefore scoped to the flash tier plus one third-party verifier.

\paragraph{Calibration scope.} We never fit a parametric calibrator; ECE/Brier/AURC are measured, not post-hoc corrected. The three-tier recalibration uses a stratified subsample whose absolute values are not comparable to full-scope tables --- only the direction of scope sensitivity is claimed.

\paragraph{Corpus time boundary.} Main-corpus collection ends around 2021; the newer-corpus experiment (\S\ref{sec:repair}) partially tests time sensitivity (entity density flat at two-year granularity, ranking consistency preserved), but extrapolating main-table precision and calibration numbers --- and any fixed threshold --- beyond that distribution demands caution. Keeping the corpus is deliberate: the critical audit's numbers share its source and split, so the ``unreliability not repaired'' comparison stands.

\paragraph{No third-party pipeline reproduction.} Following the reproducibility audit (5 of 12 systems re-runnable; F1 drifts of 0.16--0.70 across protocols), we re-execute no third-party pipeline as a same-scope baseline; cross-system numbers are quoted only from the critical audit under an explicitly different scope and never placed in one table with ours --- our frozen one-click reproduction addresses the missing baseline at the protocol level rather than the system level.

\paragraph{Preregistered hypothesis register status.} H1--H3 completed as reported (\S\ref{sec:failures}--\S\ref{sec:signals}); H4 completed (\S\ref{sec:repair}); H5 completed with scope limits (cost anchors archived, measured human hours future work); H6 completed (\S\ref{sec:counterfactual}). Human-efficiency measurement of the analyst triage workflow (as opposed to simulated review budgets) and an independent held-out calibration set remain unexecuted; they are registered as pending, not claimed. Remaining gold-bearing generalization candidates (a STIX 2.1 structured set, CTI-HAL, CyberNER, RAGRecon) are queued for a later stage. A channel interruption during the revision round was fully recovered and disclosed (all batches completed: counterfactuals 1,080/1,080 per model; verification 1,221+1,200 calls with 0 failures; flagship 822/822 with 0 failures); call-level logs are in the repository.

\paragraph*{Data availability.} All frozen batch outputs, evaluation code, annotation artifacts (the freeze JSON files with per-item labels and adjudication logs), and the derived gold are released through an anonymized repository for review, public upon acceptance; the report corpus is Zenodo deposit 7357568 (MIT) and the 2022--2025 corpus is the public Hugging Face dataset \emph{apt-cti-reports} (MIT); redistribution of model outputs is governed by vendor terms.

%% file: sections/appendix.tex
\section{APT Alias Equivalence Classes}\label{app:alias}
The dataset's 86 actor groups contain 26 multi-homing aliases. Single-value normalization misjudges Winnti Group vs.\ the gold APT41 as an FP. Bidirectional normalization via equivalence classes (a non-empty pred/gold intersection counts as a hit) raises \dsm{} APT micro-P from 0.39 to 0.57 and macro-F1 by $+0.15$. The correction table and all equivalence classes ship with the open-source code, covered by self-check assertions.

\section{EHR Sentence-Splitting Repair and Value/Sentence Misalignment}\label{app:ehr}
Splitting sentences per paragraph lets PDF line breaks destroy sentences spanning lines, deflating EHR to 0.046; switching to character-stream splitting (all line breaks folded to spaces, then split on sentence punctuation) yields 0.220 (30-report probe scope). The dominant cause of the remaining failures: the value does not exist in the sentence (the model outputs the standard name ``Spearphishing Link'' while the evidence sentence reads ``malicious hyperlink in the email attachment''). The three thresholds (0.95/0.90/0.75) are exactly the sensitivity scale of this misalignment.

\begin{table*}[!t]
\centering
\small
\begin{tabular}{lrrrr}
\toprule
Batch & Reports & Used calls & Input tok & Output tok \\
\midrule
\glm{} train & 222 & 2,202 & 17.10M & 2.99M \\
\glm{} test & 93 & 822 & 5.64M & 3.04M \\
\dsm{} train & 222 & 2,202 & 17.24M & 5.81M \\
\dsm{} test & 93 & 822 & 5.64M & 4.31M \\
ext main run (30$\times$2) & 60 & 258 & 2.05M & 1.25M \\
ext counterfactual (15$\times$2$\times$2) & 60 & 60 & 0.46M & 0.29M \\
\bottomrule
\end{tabular}
\caption{Usage and reproducibility metadata. Per-report call metadata (models/gateways/usage\_by\_gateway) is archived per batch; deterministic task ids guarantee idempotent resume without double billing; ext batches are physically separated under their own task-id prefix. Revision-round additions: relation-level counterfactuals (1,080 planned calls each for \dsm{} and \glm{}), verification baselines (\dsm{} 1,221 / \glm{} 1,200 calls), and the flagship control (822 calls, list-price $\approx$ \$28.5).}
\label{tab:usage}
\end{table*}

\section{AttackVector Scope Correspondence}\label{app:avscope}
The critical audit's AV = ``the technique exploiting the CVE'' ($\leq$4 per campaign; gold records 170 attack-vector values); this paper's AV = initial-access technique names (gold 395 entries / 13 distinct values). The conservative synonym mapping (e.g., targeted emails $\to$ spearphishing link) prefers missing a synonym over admitting a wrong one. AV numbers under the two scopes must not be compared in one table; the full mapping table ships with the code.

\begin{table*}[!t]
\centering
\small
\begin{tabular}{llcccc}
\toprule
Model & Type & vECE & sECE & vGain & sGain \\
\midrule
\glm{} train & APT & 0.534 & 0.259 & +0.328 & +0.298 \\
 & CVE & 0.851 & 0.165 & +0.041 & +0.063 \\
 & AV & 0.694 & 0.244 & +0.120 & +0.151 \\
\dsm{} train & APT & 0.652 & 0.253 & +0.168 & +0.405 \\
 & CVE & 0.938 & 0.158 & +0.022 & +0.055 \\
 & AV & 0.675 & 0.329 & +0.018 & +0.221 \\
\bottomrule
\end{tabular}
\caption{Train-split calibration (full). CVE verbalized ECE train 0.94 vs.\ test 0.837 (\glm{}) / 0.877 (\dsm{}): caused by the higher FP density and lower correctness of train-split CVEs; the direction is consistent.}
\label{tab:traincalib}
\end{table*}

\section{The 7 No-Signal Alignment Reports}\label{app:nosignal}
APT38\_\_2017-10 (second\_pass\_exact, high confidence); Gamaredon Group $\times$5 (slug channel --- the reports are written under the operation codename ``Operation Armageddon''); Lazarus Group\_\_2013-03 (slug, ``Operation Troy''). All alignments are kept as direct evidence of task difficulty --- ``reports are written under operation codenames and the APT name never appears'' --- and removing them shifts macro-F1 by less than 0.02 (sensitivity check).

\begin{table}[!ht]
\centering
\small
\captionsetup{justification=raggedright}
\begin{tabular}{rll}
\toprule
\# & Campaign & Blocked at \\
\midrule
1 & APT41\_\_2016-07 & rag / cot \\
2 & APT41\_\_2017-07 & rag / cot / rag\_cot \\
3 & APT41\_\_2018-07 & rag / cot \\
4 & CopyKittens\_\_2017-03 & rag / cot \\
5 & CopyKittens\_\_2017-04 & rag / cot \\
6 & Deep Panda\_\_2014-04 & cot \\
7 & Deep Panda\_\_2015-03 & rag \\
8 & Rancor\_\_2017-12 & rag / rag\_cot \\
9 & Tropic Trooper\_\_2012-02 & rag / rag\_cot \\
10 & Tropic Trooper\_\_2018-01 & rag / cot / rag\_cot \\
11 & Winnti Group\_\_2016-10 & rag \\
12 & APT18\_\_2015-06 & cot \\
13 & APT19\_\_2017-06 & cot \\
14 & Lazarus Group\_\_2014-11 & cot \\
15 & Taidoor\_\_2012-09 & cot \\
\bottomrule
\end{tabular}
\caption{Content-filtering gap in the ablations (15 reports, one vendor channel; \texttt{finish\_reason} = \texttt{sensitive}/\texttt{content\_filter}). Neither the \dsm{} channel nor the \glm{} main-run channel saw any blocking; ablation figures use the joint 84-report scope.}
\label{tab:contentfilter}
\end{table}

\section{Ablation Protocol Details}\label{app:ablproto}
\textbf{+RAG}: the report's full text is chunked at 1{,}000 characters with 100-character overlap; the current segment serves as the BM25 query, the top-6 chunks excluding those overlapping the current segment are concatenated as context --- in-report retrieval with no external knowledge (fending off knowledge-cutoff and data-leakage objections). \textbf{+CoT}: the prompt is extended with ``first output \texttt{Reasoning:} bullet points ($\leq$5, citing supporting sentences), then output JSON as the last content''; the parser performs multi-start brace balancing. \textbf{One extraction per configuration} (T$=$0, no SC sampling): the ablation measures single-extraction quality only. \textbf{Task isolation}: ablation task ids are derived independently of the main run, with no mutual overwriting.

\begin{table*}[!t]
\centering
\scriptsize
\setlength{\tabcolsep}{3pt}
\begin{tabular}{llccccc}
\toprule
Model & Config & APT macro-F1 ($\Delta$) & CVE ($\Delta$) & AV ($\Delta$) & mean out tok \\
\midrule
\dsm{} & base@90 & 0.679 [.603,.747] & 0.219 [.156,.286] & 0.393 [.316,.474] & 464 \\
\dsm{} & rag\_k3 & 0.711 (+0.032) [.631,.775] & 0.226 (+0.007) & 0.343 ($-$0.051) [.262,.425] & 543 \\
\dsm{} & rag\_k6 & 0.704 (+0.025) [.626,.777] & 0.226 (+0.007) & 0.378 ($-$0.016) [.292,.465] & 566 \\
\dsm{} & rag\_k12 & 0.703 (+0.025) & 0.227 (+0.008) & 0.383 ($-$0.011) & --- \\
\dsm{} & cot3 & 0.691 (+0.013) [.613,.770] & 0.233 (+0.014) & 0.311 ($-$0.083) [.239,.396] & 430 \\
\dsm{} & cot5 & 0.678 (+0.000) [.597,.762] & 0.220 (+0.001) & 0.281 ($-$0.112) [.210,.356] & 481 \\
\dsm{} & cot10 & 0.681 (+0.002) [.600,.754] & 0.226 (+0.008) & 0.259 ($-$0.134) [.186,.334] & 520 \\
\dsm{} & schema\_cap & 0.724 (+0.046) [.643,.794] & 0.226 (+0.007) & 0.349 ($-$0.045) [.260,.434] & 241 \\
\dsm{} & schema\_main & \textbf{0.789 (+0.111)} [.710,.858] & 0.234 (+0.015) & 0.328 ($-$0.065) [.240,.420] & 346 \\
\dsm{} & schema\_noev & 0.680 (+0.001) [.599,.755] & 0.222 (+0.004) & 0.348 ($-$0.045) [.265,.424] & 263 \\
\glm{} & base@90 & 0.637 [.567,.704] & 0.221 [.157,.289] & 0.308 [.249,.373] & --- \\
\glm{} & rag\_k3 & 0.719 (+0.083) [.649,.789] & 0.219 ($-$0.002) & 0.344 (+0.036) [.267,.433] & 3,103 \\
\glm{} & rag\_k12 & 0.683 (+0.046) [.604,.751] & 0.221 (+0.001) & 0.389 (+0.081) [.299,.472] & 3,018 \\
\glm{} & cot3 & 0.698 (+0.061) [.623,.771] & 0.217 ($-$0.004) & 0.365 (+0.057) [.283,.451] & 3,002 \\
\glm{} & cot5 & 0.762 (+0.125) [.689,.840] & 0.207 ($-$0.014) & 0.372 (+0.064) [.293,.455] & 2,597 \\
\glm{} & cot10 & 0.712 (+0.075) [.637,.778] & 0.217 ($-$0.004) & 0.352 (+0.044) [.275,.436] & 3,294 \\
\glm{} & schema\_cap & 0.746 (+0.110) [.680,.817] & 0.240 (+0.020) & 0.366 (+0.058) [.283,.452] & 2,273 \\
\glm{} & schema\_main & \textbf{0.803 (+0.167)} [.736,.869] & 0.234 (+0.013) & 0.374 (+0.066) [.287,.461] & 2,756 \\
\glm{} & schema\_noev & 0.720 (+0.083) [.647,.789] & 0.222 (+0.002) & 0.409 (+0.101) [.331,.492] & 2,126 \\
\bottomrule
\end{tabular}
\caption{Second-stage ablation: $\Delta$ macro-F1 / absolute values vs.\ the same model's base (@90-report intersection; 95\% bootstrap CI where shown; omitted CIs are of the same magnitude as adjacent same-model rows, full CIs in \texttt{ablation\_summary.json}). Token column is \dsm{}-side measured; \glm{}-side thinking cannot be disabled (output tokens 5--6$\times$). \texttt{schema\_noev} reports only PRF. Flagship scope: ablations were not run on \gpt{} (repair-path conclusions scoped to the flash tier).}
\label{tab:secondabl}
\end{table*}

\section{Second-Stage Ablation: Three Knob Readings}\label{app:secondabl}
\textbf{top-k}: F1 is insensitive to $k$ while input tokens are strictly monotone (10.7k$\to$12.6k) --- the retrieval gain does not come from widening the window; the only non-flat response is \glm{}-side (APT falls 0.719$\to$0.683 as AV rises 0.344$\to$0.389 with $k$), consistent with over-extraction dominating FPs. \textbf{CoT length}: \dsm{} is length-insensitive with strictly monotone tokens (pick the shortest); \glm{} peaks numerically at $\leq$5 (+0.125) with overlapping CIs --- evidence does not support ``longer reasoning is better''. \textbf{Schema variants} give the two-sided test: \texttt{schema\_main} pushes APT macro-F1 sharply (\dsm{} +0.111, \glm{} +0.167, CIs non-overlapping with base) yet raises CVE on neither model --- \textbf{primary-entity alignment repairs actor-type over-extraction but cannot repair the CVE scope mismatch}; \texttt{schema\_cap} (hard cap of 3) raises \dsm{} CVE micro-P from 0.060 to 0.381 with flat macro-F1 (suppresses quantity, not misalignment); \texttt{schema\_noev} leaves F1 at base with the lowest tokens.

\begin{table*}[!t]
\centering
\small
\begin{tabular}{lll}
\toprule
Hypothesis & Status & Evidence \\
\midrule
H1 three-tier annotation & completed \& frozen & \S\ref{sec:failures}, Table~\ref{tab:threetier} \\
H2 SC increment (ctrl.\ text freq.) & completed & \S\ref{sec:signals} \\
H3 review budget + audit-label recalib. & completed & \S\ref{sec:signals} \\
H4 gold-bearing extrapolation & completed & \S\ref{sec:repair} \\
H5 review simulation vs.\ human effort & compl.\ w/ scope limits & Limitations \\
H6 third-party calibrated verifier & completed & \S\ref{sec:counterfactual} \\
\bottomrule
\end{tabular}
\caption{Preregistered hypothesis register (H1--H6), final status. All acceptance lines and mechanisms were registered before evaluation; failing a registered bar must be restated as a negative result.}
\label{tab:hregister}
\end{table*}